\documentclass{article}
\usepackage{ijcai26}

\usepackage{times}
\usepackage{soul}
\usepackage{url}
\usepackage[hidelinks]{hyperref}
\usepackage[utf8]{inputenc}
\usepackage[small]{caption}
\usepackage{graphicx}
\usepackage{amsthm}
\usepackage{booktabs}
\usepackage[switch]{lineno}
\usepackage{amssymb}
\usepackage{amsmath} 
\usepackage{multirow}

\usepackage{xcolor} 

\usepackage{algorithm} 

\usepackage{algpseudocode} 

\usepackage{bbding}

\title{Nipping the Drift in the Bud: Retrospective Rectification for Robust \\Vision-Language Navigation}

\author{
Gang He$^1$
\and Zhenyang Liu$^1$
\and Kepeng Xu$^1$
\and Li Xu$^1$
\and Tong Qiao$^1$ \\ 
Wenxin Yu$^2$
\and Chang Wu$^1$
\and Weiying Xie$^1$ \\
\affiliations
$^1$Xidian University\\
$^2$Southwest University of Science and Technology\\
\emails
kepengxu11@gmail.com
}

\begin{document}

\maketitle

\begin{abstract}
Vision-Language Navigation (VLN) requires embodied agents to interpret natural language instructions and navigate through complex continuous 3D environments. However, the dominant imitation learning paradigm suffers from exposure bias, where minor deviations during inference lead to compounding errors. While DAgger-style approaches attempt to mitigate this by correcting error states, we identify a critical limitation: Instruction-State Misalignment. Forcing an agent to learn recovery actions from off-track states often creates supervision signals that semantically conflict with the original instruction. In response to these challenges, we introduce \textit{BudVLN}, an online framework that learns from on-policy rollouts by constructing supervision to match the current state distribution. \textit{BudVLN} performs retrospective rectification via counterfactual re-anchoring and decision-conditioned supervision synthesis, using a geodesic oracle to synthesize corrective trajectories that originate from valid historical states, ensuring semantic consistency. Experiments on the standard R2R-CE and RxR-CE benchmarks demonstrate that \textit{BudVLN} consistently mitigates distribution shift and achieves state-of-the-art performance in both Success Rate and SPL.
\end{abstract}

\section{Introduction}
\label{sec:intro}

\begin{figure*}[t] 
    \centering
    \includegraphics[width=2.0\columnwidth]{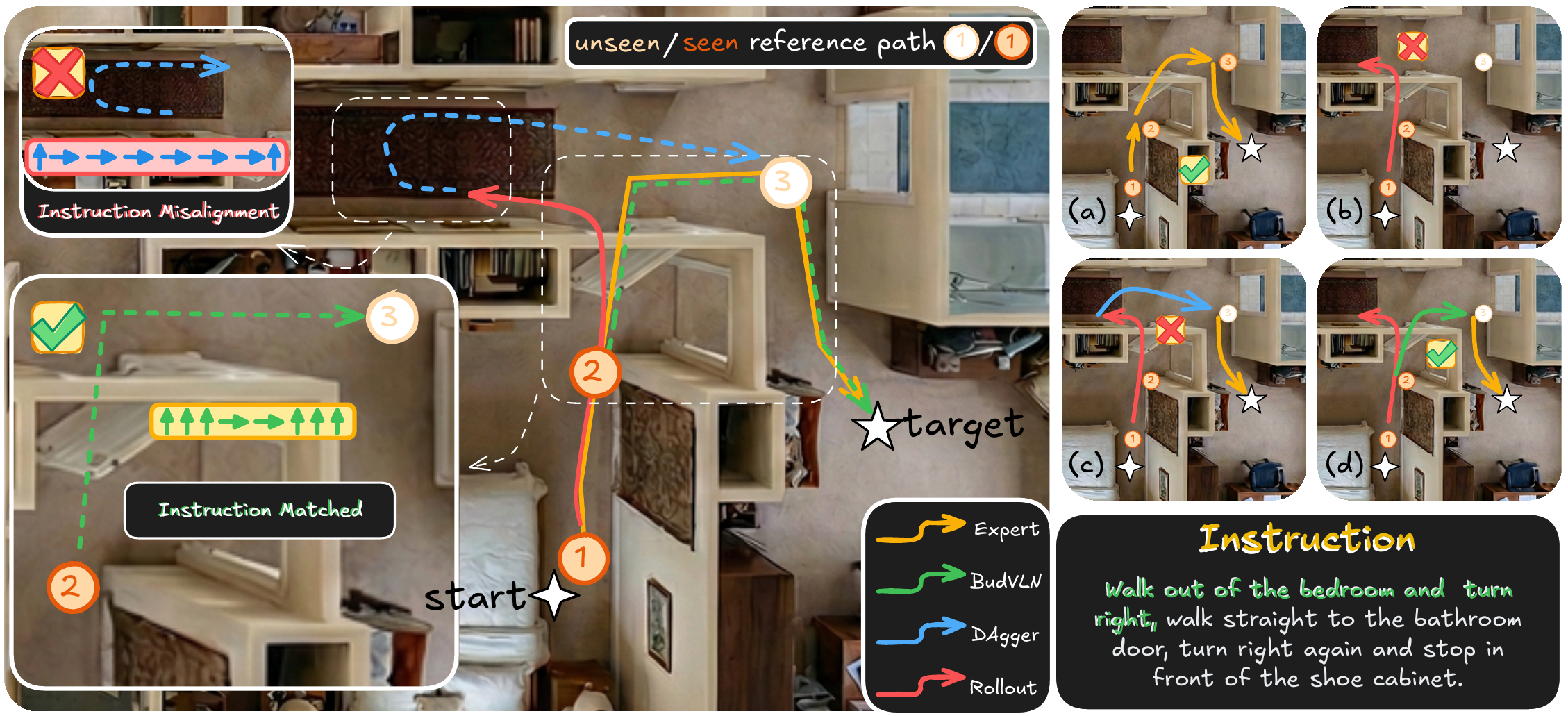} 
    \caption{\textbf{Illustration of instruction-state misalignment.} 
    The yellow line in (a) depicts a static expert demonstration. 
    When the online rollout deviates \textbf{due to navigational uncertainty} (red line in (b)), standard DAgger forces a recovery from the error state. 
    However, the required \textbf{backward-correcting} actions (e.g., turning back (blue line in (c))) fail to establish a semantic connection with the instruction \textit{walk straight}, leading to further grounding confusion. 
    In contrast, our BudVLN employs retrospective rectification (green line in (d)): it re-anchors to the latest progressed point on the reference path to synthesize a \textbf{forward-looking} demonstration that remains strictly aligned with the natural language instruction.}
    \label{fig:teaser}
\end{figure*}
Vision-Language Navigation (VLN) requires an embodied agent to navigate through photorealistic 3D environments by following natural language instructions \cite{anderson2018vision,ku2020room,krantz2020beyond}. 
While recent advances in large multimodal models have significantly improved agent capabilities \cite{wei2025streamvln,zhang2024navid,cheng2024navila}, the dominant training paradigm remains Imitation Learning (IL) via Teacher Forcing on static expert demonstrations. 
This paradigm suffers from a fundamental limitation known as \textit{exposure bias}: during training, the agent mimics actions conditioned on ground-truth states, but at inference, it must act upon its own policy-induced states \cite{ross2011reduction,tan2019learning}. 
Even minor perturbations can drive the agent into Out-Of-Distribution (OOD) regions, where the lack of valid supervision leads to compounding errors and catastrophic failure.

To mitigate this distribution shift, existing literature primarily adopts two methodological paradigms: Reinforcement Learning (RL) \cite{tan2019learning,qi2025vlnr1,zhang2025activevln} and Interactive Imitation Learning (e.g., DAgger) \cite{ross2011reduction}.
RL theoretically allows the agent to update its policy based on its own induced distribution via online exploration.
However, for samples where the model exhibits high uncertainty, RL suffers from extremely low exploration efficiency \cite{zhang2025activevln}.
Due to the long-horizon nature of VLN and the vast state-action space, the spontaneous discovery of correct paths becomes computationally intractable when relying solely on sparse, hand-crafted reward signals, which often fail to align effectively with the fine-grained semantics of the instructions.
Conversely, DAgger-style approaches provide explicit guidance for error states, effectively mitigating exposure bias to some degree. However, we argue that this trajectory-level robustness is often achieved at the cost of semantic consistency. Standard DAgger inherently introduces \textit{Instruction-State Misalignment} (Figure \ref{fig:teaser}).
For instance, correcting a large deviation often requires demonstrating a path that first returns to the reference trajectory.
Such recovery maneuvers compel the model to learn behaviors that are semantically inconsistent with the original instruction, thereby severing the alignment between the linguistic directives and the optimal actions.

In this paper, we propose BudVLN, a robust framework designed to resolve these challenges through a unified online retrospective rectification.
To fundamentally mitigate exposure bias, BudVLN operates on a continuous online loop where the agent updates its policy based on-policy rollouts.
Unlike static imitation, this ensures the agent learns directly from its own induced observation distribution, bridging the gap between training and inference.
To address the inefficiency of RL on hard samples, we establish a dynamic synergy between exploration and supervision: 
we utilize Group Relative Policy Optimization (GRPO) to refine proficient samples, allowing the agent to autonomously explore more precise and instruction-aligned paths; 
concurrently, for hard samples where stochastic exploration fails, we trigger Supervised Fine-Tuning (SFT) to provide explicit, high-quality guidance.
Crucially, to overcome the Instruction-State Misalignment of standard DAgger, our SFT component employs a Retrospective Rectification mechanism.
Instead of teaching the agent how to recover from an error state (which often conflicts with the instruction), we teach it how to avoid entering the error state from the outset.
By re-anchoring to the last valid progress point and synthesizing a semantically consistent demonstration, we ensure the synthesized supervision maintains strict consistency between the visual history and the linguistic directives, eliminating the adversarial signals found in traditional recovery behaviors.

Our main contributions are summarized as follows:
\begin{itemize}
    \item \textbf{Synergistic Online Training Framework:} We propose a unified online training framework that effectively mitigates exposure bias while resolving the exploration inefficiency of pure RL. By dynamically synergizing Group Relative Policy Optimization (GRPO) for proficient samples and Supervised Fine-Tuning (SFT) for hard samples, our approach enables the agent to continuously improve via self-exploration while receiving timely guidance where it lacks proficiency.
    
    \item \textbf{Retrospective Rectification Mechanism:} We identify and address the \textit{Instruction-State Misalignment} problem inherent in standard DAgger-style learning. Instead of forcing the agent to learn semantically ambiguous recovery actions from error states, our Retrospective Rectification re-anchors the agent to valid historical states to synthesize corrective demonstrations, ensuring strict alignment between visual observations and linguistic instructions.
    
    \item \textbf{SOTA Performance:} We demonstrate through extensive experiments that BudVLN achieves state-of-the-art performance on standard continuous VLN benchmarks, yielding significant improvements in both Success Rate (SR) and SPL. Furthermore, our selective update strategy proves highly sample-efficient, achieving these gains with only $\sim$25\% of the training cost required by standard DAgger pipelines, offering a scalable paradigm for training embodied agents.
\end{itemize}

\section{Related Work}

\subsection{Vision-Language Navigation Agents}
The evolution of Vision-Language Navigation agents has largely mirrored the advancements in deep learning architectures. 
Early approaches relied on LSTM-based sequence-to-sequence models grounded in discrete environments \cite{anderson2018vision,tan2019learning}. 
With the advent of Transformers, graph-based methods like HAMT \cite{chen2021hamt} and DUET \cite{chen2022duet} achieved significant performance by modeling global topological maps. 
However, deploying these agents in continuous environments (VLN-CE) \cite{krantz2020beyond} introduces challenges related to low-level control and infinite state spaces. 
While methods like ETPNav \cite{an2024etpnav} address this via online topological mapping, they often require complex, hand-crafted navigation heuristics.

Recent works have begun to leverage the reasoning capabilities of Large Language Models (LLMs) and Vision-Language Models (VLMs) for embodied control. 
NaVid \cite{zhang2024navid} and NaVILA \cite{cheng2024navila} demonstrate that video-based foundation models can directly predict low-level actions from raw visual streams without explicit map building. 
Similarly, StreamVLN \cite{wei2025streamvln} proposes a streaming architecture with slow-fast context modeling to handle long-horizon visual history efficiently. 
However, most of these agents are still optimized via static imitation learning, which struggles under on-policy deviations at test time. 
In contrast, we focus on a robust online training framework that explicitly targets distribution shift in VLM control policies.

\subsection{Imitation Learning and Supervision Construction}
Imitation Learning (IL), specifically Behavior Cloning (BC) or Teacher-Forcing, remains the dominant training paradigm for VLN \cite{wei2025streamvln,zhang2024navid}. 
It is well-established that BC suffers from exposure bias \cite{ross2011reduction}, where the agent fails to recover once it drifts from the expert state distribution. 
To mitigate this, DAgger (Dataset Aggregation) \cite{ross2011reduction} and its variants iteratively collect on-policy data and query an expert for ground-truth actions. 
In the VLN domain, approaches like CorrectNav \cite{yu2025correctnav} and others \cite{li2022envedit} attempt to correct the agent by providing expert paths starting directly from the error states. 
While effective for reducing trajectory errors, we argue that this standard DAgger practice introduces \textit{Instruction-State Misalignment}. 
Specifically, the geometric maneuvers required to recover from large deviations often explicitly contradict the directional semantics of the original instruction (e.g., turning back vs. walking straight), thereby generating adversarial supervision signals that confuse the agent's language grounding.
Unlike these methods, BudVLN addresses instruction-state misalignment by retrospectively synthesizing supervision anchored to valid historical states. This ensures the agent learns from semantically consistent trajectories rather than ambiguous recovery actions.

\subsection{Reinforcement Learning in Embodied AI}
Reinforcement Learning (RL) introduces an online interaction paradigm, allowing agents to improve through trial and error beyond static datasets. 
Prior works like ActiveVLN \cite{zhang2025activevln} utilize RL to encourage active exploration. 
However, relying solely on RL with sparse rewards faces significant challenges in the high-dimensional, long-horizon VLN task. 
Especially for unsolved instructions, the vast action space renders stochastic exploration computationally prohibitive for discovering successful trajectories.
Consequently, pure RL often struggles to acquire valid learning signals for these hard samples.

In this work, we leverage RL not merely as an optimizer, but as an online framework to enable dynamic supervision construction. 
We adopt Group Relative Policy Optimization (GRPO) \cite{shao2024deepseekmath,guo2025deepseek,qi2025vlnr1} to drive this online loop. 
Unlike traditional architectures, our framework establishes a synergistic mechanism: 
GRPO focuses on efficiency optimization by reinforcing successful trajectories to discover optimal, robust paths; 
concurrently, our Retrospective Rectification (SFT) handles failure prevention by explicitly demonstrating how to avert errors via historical re-anchoring. 
This hybrid approach complements the open-ended exploration of RL with targeted guidance, retaining the benefits of online policy improvement while ensuring effective learning from failures.

\section{Methodology}
\begin{figure*}[t]
    \centering
    \includegraphics[width=\textwidth]{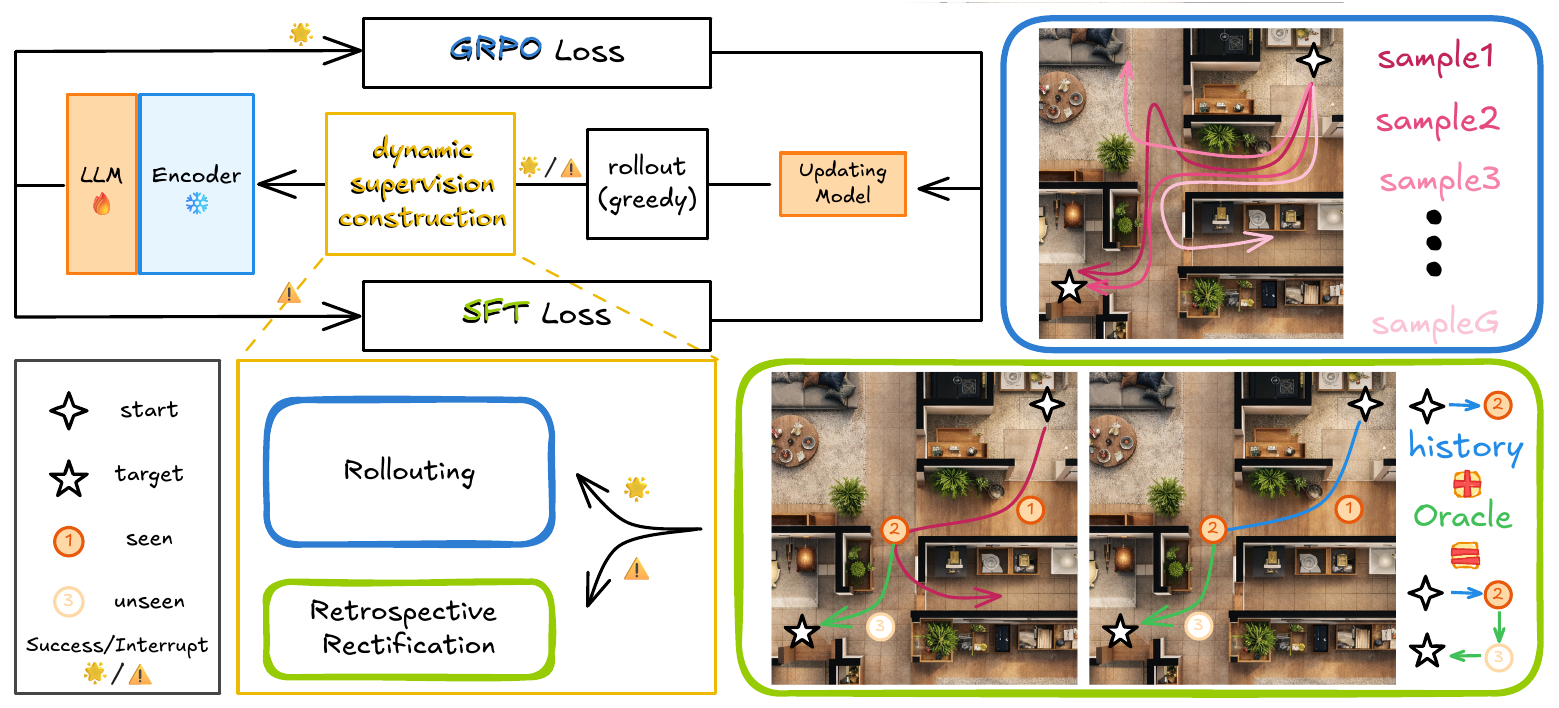}
    \caption{\textbf{Overview of the BudVLN training framework.} 
    The framework employs an \textit{Adaptive Mutual Exclusion Strategy} to harmonize exploration and supervision. 
    For a given instruction, a greedy probe first evaluates the agent's proficiency. 
    When proficient, the framework routes to the optimality seeking pathway, employing GRPO to reinforce high-SPL behaviors via diverse sampling. 
    Conversely, upon failure, it triggers the rectification pathway, which reverts to a valid historical state to synthesize alignment-preserving supervision. 
    Gradients are back-propagated exclusively from the active pathway to ensure stable policy updates.} 
    \label{fig:framework}
\end{figure*}
\subsection{Problem Formulation}
We formulate VLN-CE~\cite{krantz2020beyond} as a Partially Observable Markov Decision Process (POMDP) defined by the tuple $\langle \mathcal{S}, \mathcal{A}, \mathcal{O}, \mathcal{T}, \mathcal{R}, I \rangle$.
At each time step $t$, the agent receives a natural language instruction $I$ and a visual observation $o_t$.
To adhere to the standard RGB-only setting~\cite{zhang2024navid}, we define $o_t$ as monocular RGB images, without assuming access to depth, odometry, or pre-built maps.
The agent maintains a history context $H_t=\{I,o_0,a_0,\ldots,o_t\}$ to capture temporal dependencies.
Based on $H_t$, the policy $\pi_\theta(a_t\mid H_t)$ predicts a discrete action token $a_t\in\mathcal{A}$ (e.g., \texttt{FORWARD}, \texttt{TURN LEFT}, \texttt{TURN RIGHT}, \texttt{STOP}), which is then executed by a low-level controller to produce continuous movements.
The objective is to maximize the expected return, which is designed to align with the SPL metric.

\subsection{BudVLN Framework Overview}
\label{sec:framework}

To address the challenges of exploration efficiency and adversarial supervision discussed in Sec. \ref{sec:intro}, we propose \textbf{BudVLN}, a unified online training framework. 
Building upon representative policy architectures in the VLN domain, BudVLN operates in iterative online rounds. 
Unlike traditional pipelines that treat exploration and correction independently, we establish an Adaptive Mutual Exclusion Strategy to dynamically bifurcate the training process based on the agent's real-time proficiency.

As illustrated in Figure~\ref{fig:framework}, for each instruction $I$ in a batch, the process proceeds as follows:

\noindent\textbf{1. Greedy Probe.} 
The agent first executes a single trajectory $\tau_{probe}$ using deterministic greedy decoding. This serves as a low-cost probe to assess the policy's current mastery of the instruction without the overhead of group sampling.

\noindent\textbf{2. Dynamic Routing.} 
We evaluate $\tau_{probe}$ against a set of failure triggers. The training objective is then conditionally determined, routing the gradient flow into one of two exclusive pathways:
\begin{description}
    \item[\textbf{Proficiency Pathway (via GRPO Loss).}] 
    If $\tau_{probe}$ is successful (i.e., no triggers activated), we classify the sample as proficient. 
    To encourage the discovery of superior paths, the agent samples $G-1$ additional stochastic trajectories. 
    We then update the policy using the \textbf{GRPO objective} (Eq. \ref{eq:grpo_loss}) to reinforce efficiency, strictly bypassing the SFT computation.
    
    \item[\textbf{Rectification Pathway (via SFT Loss).}] 
    If $\tau_{probe}$ triggers a failure mode, we classify the sample as hard. 
    To avoid inefficient exploration, we discard stochastic rollouts and trigger the Retrospective Rectification mechanism. 
    The policy is updated using the \textbf{Weighted SFT objective} (Eq. \ref{eq:rect_loss}) based on the synthesized supervision, skipping the GRPO update.
\end{description}

This mutual exclusion ensures that the gradients are either driven by relative efficiency optimization (for mastered skills) or direct error correction, but never both simultaneously for the same sample.

\subsection{Optimality-Seeking Exploration via GRPO}
\label{sec:grpo}
When the greedy probe confirms the agent's proficiency (i.e., successfully reaching the goal), the learning objective shifts from basic completion to optimality-seeking---specifically, finding shorter, more efficient paths.
We adopt Group Relative Policy Optimization (GRPO)~\cite{shao2024deepseekmath} to reinforce such superior strategies.

\noindent\textbf{Group Sampling and Reward.}
For a proficient instruction $I$, we sample a group of $G$ trajectories $\boldsymbol{\tau}=\{\tau_1,\dots,\tau_G\}$, including the greedy probe and $G-1$ stochastic rollouts with temperature $0.4$.
Each trajectory $\tau_i$ yields a scalar return $R(\tau_i)$.
To explicitly encourage efficiency, our reward incorporates the SPL (Success weighted by Path Length) metric:
\begin{equation}
\label{eq:reward}
R(\tau_i)=\mathbb{I}(\text{success})\cdot\big(C_{\text{succ}}+\lambda\cdot \text{SPL}_i\big)-C_{\text{dist}}\cdot d_{\text{remain}},
\end{equation}
where $\mathbb{I}(\text{success})$ indicates task completion, $C_{\text{succ}}=2.0$ is the base success bonus, $\lambda$ weights the $\text{SPL}_i$ term, and $C_{\text{dist}}$ scales the penalty for the remaining geodesic distance to the goal $d_{\text{remain}}$.

\noindent\textbf{Relative Advantage Estimation.}
GRPO avoids training an explicit value critic and instead estimates the baseline using group statistics.
Given the sampled trajectories $\{\tau_1,\dots,\tau_G\}$ and their returns $\{R(\tau_1),\dots,R(\tau_G)\}$, we compute the normalized group-relative advantage for each trajectory:
\begin{equation}
\label{eq:advantage}
A_i=\frac{R(\tau_i)-\mathrm{mean}\!\left(\{R(\tau_1),\dots,R(\tau_G)\}\right)}
{\mathrm{std}\!\left(\{R(\tau_1),\dots,R(\tau_G)\}\right)+\epsilon}.
\end{equation}

\noindent\textbf{GRPO Objective with KL Regularization.}
Let $\tau_i=\{(H_{i,t}, a_{i,t})\}_{t=1}^{T_i}$ denote the $i$-th trajectory, where $H_{i,t}$ is the agent history context (instruction and observations/actions up to step $t$), and $a_{i,t}$ is the action taken at step $t$.
Define the probability ratio
\begin{equation}
\label{eq:ratio}
\rho_{i,t}=\frac{\pi_\theta(a_{i,t}\mid H_{i,t})}{\pi_{\theta_{\mathrm{old}}}(a_{i,t}\mid H_{i,t})}.
\end{equation}
We optimize the following clipped surrogate objective while constraining the policy to stay close to a reference policy $\pi_{\mathrm{ref}}$ via a KL penalty:
\begin{equation}
\label{eq:grpo_loss}
\begin{aligned}
\mathcal{L}_{\text{GRPO}}(\theta)
= \mathbb{E}\Bigg[
& \frac{1}{G}\sum_{i=1}^{G}\frac{1}{T_i}\sum_{t=1}^{T_i}
\Big(
\min\big(\rho_{i,t}A_i,\;
\mathrm{clip}(\rho_{i,t},1-\epsilon,1+\epsilon)A_i\big) \\
& \qquad - \beta\,\mathbb{D}_{\mathrm{KL}}\!\big(
\pi_\theta(\cdot\mid H_{i,t}) \;\|\;
\pi_{\mathrm{ref}}(\cdot\mid H_{i,t})
\big)
\Big)
\Bigg].
\end{aligned}
\end{equation}

where $\epsilon$ is the clipping parameter, $\beta$ controls the strength of the KL regularization, and the KL term is computed between the action distributions conditioned on the same history $H_{i,t}$.

\subsection{Retrospective Rectification (SFT)}
\label{sec:rectification}
For hard samples where the greedy probe fails, stochastic exploration is inefficient. 
We employ Retrospective Rectification to construct valid, alignment-preserving supervision from the failed attempt.

\noindent \textbf{Oracle Demonstration Triggers.}
We define four specific triggers to detect failures during the greedy probe:
\begin{enumerate}
    \item Off-track: Deviation from the reference path $>3.0$m or heading error $>120^{\circ}$;
    \item Progress Stall: No new reference waypoint visited for 60 steps;
    \item Premature Stop: Stopping $>3.0$m from the goal;
    \item Forced Stop: Failing to stop within the goal zone after a grace period.
\end{enumerate}

\noindent \textbf{History-Aware Rollback.}
Upon triggering, we perform a History-Aware Rollback to synthesize a corrective demonstration:
\begin{enumerate}
    \item Anchor Identification: We identify the best waypoint state $s_{anchor}$ achieved in the failed trajectory, defined as the state corresponding to the furthest reference path on the ground-truth graph that was successfully visited before the failure.
    \item Valid Context Retention: We extract the policy's observation history $H_{anchor}$ from the start up to $s_{anchor}$. Crucially, we retain this valid prefix (generated by the agent itself) while discarding the subsequent erroneous actions.
    \item Oracle Completion: We reset the simulator state to $s_{anchor}$. An oracle planner then generates the optimal action sequence $\mathbf{a}^*_{corr}$ from $s_{anchor}$ to the goal.
\end{enumerate}

This process yields a supervision pair $(H_{anchor}, \mathbf{a}^*_{corr})$. It effectively teaches the agent: Given that you have successfully reached $s_{anchor}$ via your own history $H_{anchor}$, here is the optimal path forward. 
This avoids the Instruction-State Misalignment of correcting from an off-track state.

\noindent \textbf{Weighted SFT Loss.}
We optimize the policy using a weighted Cross-Entropy loss on the oracle tokens:
\begin{equation}
    \label{eq:rect_loss} 
    \mathcal{L}_{\text{Rect}}(\theta) = - \sum_{t=1}^{|\mathbf{a}^*_{corr}|} w_t \log \pi_\theta(a^*_{t} | H_{anchor}, a^*_{<t})
\end{equation}
To emphasize immediate correction, the weights $w_t$ decay exponentially from the rollback point ($w_0=1.0$) to the end of the trajectory, prioritizing the critical decisions required to resume correct navigation.

\subsection{Optimization and Algorithm}
\label{sec:optimization}

\begin{algorithm}[t]
\captionsetup{justification=centering}
\caption{\textbf{GRO}: \\Greedy-Routed Optimization for BudVLN}
    \label{alg:GRO_Algorithm}

\begin{algorithmic}[1]
    \Statex \textbf{Input:} Dataset $\mathcal{D}$, Planner $\mathcal{P}^*$
    \Statex \textbf{Hyperparam:} Policy $\pi_\theta$, Group size $G$, Rate $\eta$
    \Statex \textbf{Output:} Optimized Policy $\pi_{\theta^*}$
    
    \vspace{0.1cm}
    \hrule
    \vspace{0.1cm}

    \State $\pi_\theta \leftarrow \text{Pretrained}(\pi_{\text{base}})$
    
    \While{not converged}
        \State $I \sim \mathcal{D}$ \Comment{Sample episode}
        \State $\tau_{\text{probe}} \leftarrow \text{Rollout}(\pi_\theta, I, \text{greedy})$ \Comment{Greedy Probe}
        
        \If{$\text{IsSuccess}(\tau_{probe})$} 
            \State $\{\tau_i\}_{i=1}^G \sim \pi_\theta(\cdot | I)$ 
            \State $A_{i} \leftarrow \displaystyle \frac{R(\tau_i) - \mu_R}{\sigma_R + \epsilon}$
            \State $g \leftarrow \displaystyle \frac{1}{G} \sum_{i=1}^G \nabla_\theta \mathcal{L}_{\text{GRPO}}(\tau_i, A_i)$ 
            \State $\theta \leftarrow \theta - \eta \cdot g$ 
            
        \Else \Comment{Dynamic Anchoring}

            \State $(s_{\text{anc}}, H_{\text{anc}}) \leftarrow \text{GetAnchorState}(\tau_{\text{probe}})$
            
            \State $\mathbf{a}^* \leftarrow \mathcal{P}^*(s_{\text{anc}}, s_{\text{anc+1}},...,\text{goal})$ 
            
            \State $g \leftarrow \nabla_\theta \mathcal{L}_{\text{SFT}}(H_{\text{anc}}, \mathbf{a}^*)$ 
            \State $\theta \leftarrow \theta - \eta \cdot g$ 
        \EndIf
    \EndWhile
\end{algorithmic}
\end{algorithm}

BudVLN employs an exclusive update strategy to stabilize training. In each optimization step, the loss is dynamically selected based on the outcome of the greedy probe $\tau_{probe}$:
\begin{equation}
    \mathcal{L}_{\text{Total}} = \mathbb{I}(\tau_{probe} \in \mathcal{S}_{succ}) \cdot \mathcal{L}_{\text{GRPO}} + \mathbb{I}(\tau_{probe} \notin \mathcal{S}_{succ}) \cdot \mathcal{L}_{\text{Rect}}
\end{equation}
where $\mathbb{I}(\cdot)$ is the indicator function, and $\mathcal{S}_{succ}$ denotes the set of trajectories that successfully reach the goal without triggering any failure modes. 
This formulation, as detailed in Algorithm \ref{alg:GRO_Algorithm} GRO, ensures that gradients are derived from relative efficiency optimization when $\tau_{probe}$ demonstrates proficiency, or from retrospective rectification when it fails.

\section{Experiments}
\label{sec:experiments}

\begin{table*}[t] 
\centering
\resizebox{0.95\textwidth}{!}{ 
\begin{tabular}{lccccccccc}
\toprule
\multirow{2}{*}{Method} & \multirow{2}{*}{Stage} & \multicolumn{4}{c}{R2R-CE Val Unseen} & \multicolumn{4}{c}{RxR-CE Val Unseen} \\
\cmidrule(lr){3-6} \cmidrule(lr){7-10}
 & & NE$\downarrow$ & OS$\uparrow$ & SR$\uparrow$ & SPL$\uparrow$ & NE$\downarrow$ & SR$\uparrow$ & SPL$\uparrow$ & nDTW$\uparrow$ \\
\midrule
Seq2Seq$^\ddagger$ \cite{krantz2020beyond} & IL & 7.77 & 37.0 & 25.0 & 22.0 & 12.10 & 13.9 & 11.9 & 30.8 \\
R2R-CMTP$^\ddagger$ \cite{chen2021topological} & IL & 7.90 & 38.0 & 26.4 & 22.7 & - & - & - & - \\
LAW$^\ddagger$ \cite{raychaudhuri2021law} & IL & 6.83 & 44.0 & 35.0 & 31.0 & 10.90 & 8.0 & 8.0 & 38.0 \\

HPN+DN$^\ddagger$ \cite{krantz2021waypoint} & IL & 6.31 & 40.0 & 36.0 & 34.0 & - & - & - & - \\
CMA$^\ddagger$ \cite{hong2022bridging} & IL & 6.20 & 52.0 & 41.0 & 36.0 & 8.76 & 26.5 & 22.1 & 47.0 \\
VLN$\circlearrowright$BERT$^\ddagger$ \cite{hong2022bridging} & IL & 5.74 & 53.0 & 44.0 & 39.0 & 8.98 & 27.0 & 22.6 & 46.7 \\
GridMM$^\ddagger$ \cite{wang2023gridmm} & IL & 5.11 & 61.0 & 49.0 & 41.0 & - & - & - & - \\
ScaleVLN$^\ddagger$ \cite{wang2023scalevln} & IL & 4.80 & - & 55.0 & 51.0 & - & - & - & - \\
ETPNav$^\ddagger$ \cite{an2024etpnav} & IL & 4.71 & 65.0 & 57.0 & 49.0 & 5.64 & 54.7 & 44.8 & 61.9 \\

\midrule
NaVid \cite{zhang2024navid} & IL & 5.47 & 49.1 & 37.4 & 35.9 & 8.41 & 34.5 & 23.8 & 21.2 \\
UniNaVid \cite{zhang2025uninavid} & IL & 5.58 & 53.3 & 47.0 & 42.7 & 6.24 & 48.7 & 40.9 & - \\
NaVILA \cite{cheng2024navila} & IL & 5.22 & 62.5 & 54.0 & 49.0 & 6.77 & 49.3 & 44.0 & 58.8 \\
StreamVLN \cite{wei2025streamvln} & IL & 4.87 & 64.3 & 57.0 & 50.5 & 6.22 & 52.9 & 46.0 & 61.9 \\

\midrule
VLN-R1 \cite{qi2025vlnr1} & IL+RL & 7.00 & 41.2 & 30.2 & 21.8 & 10.4 & 22.3 & 17.5 & - \\
ActiveVLN \cite{zhang2025activevln} & IL+RL & 5.34 & 62.1 & 52.9 & 45.7 & 5.84 & 50.7 & 41.2 & 58.1 \\
\textbf{BudVLN (Ours)} & \textbf{IL+RL} & \textbf{4.74} & \textbf{65.6} & \textbf{57.6} & \textbf{51.1} & \textbf{5.79} & \textbf{56.1} & \textbf{46.6} & \textbf{63.2} \\
\bottomrule
\end{tabular}
}
\caption{Comparison with state-of-the-art methods on R2R-CE and RxR-CE Val Unseen splits. 
Methods are categorized by learning paradigms: Imitation Learning (IL) and hybrid IL+RL. 
$\ddagger$ indicates methods utilizing auxiliary inputs (e.g., depth, maps, panoramic views).}
\label{tab:combined_results}
\end{table*}

\subsection{Experimental Setup}

\noindent\textbf{Datasets.} 
We evaluate BudVLN on two standard benchmarks (e.g., MP3D~\cite{chang2017matterport3d}) for Vision-Language Navigation in Continuous Environments (VLN-CE): R2R-CE \cite{krantz2020beyond} and RxR-CE \cite{ku2020room}. 
R2R-CE contains 5.6k trajectories with English instructions, while RxR-CE is a larger, multilingual dataset with more complex paths.
We perform joint training on the union of R2R-CE and RxR-CE training splits. All experiments are conducted using standard supervision signals without external datasets (e.g., HM3D~\cite{ramakrishnan2021habitat}) or synthetic instructions (e.g., EnvDrop).
We report performance on the respective val unseen splits of each benchmark.

\noindent\textbf{Evaluation Metrics.} 
Following standard protocols~\cite{anderson2018vision}, we employ four primary metrics: 
(1) Success Rate (SR): The percentage of trajectories stopping within 3.0m of the goal; 
(2) Success weighted by Path Length (SPL): SR penalized by the path length ratio, serving as the primary indicator of navigation efficiency and optimality; 
(3) Oracle Success Rate (OSR): The success rate if the agent stops at the closest point to the goal along its path; 
(4) Navigation Error (NE): The average geodesic distance from the stopping point to the goal.

\noindent\textbf{Implementation Details.} 
For the pre-training strategy, we strictly follow the setup of StreamVLN~\cite{wei2025streamvln}.
In the BudVLN fine-tuning phase, we employ a group size of $G=4$ with a sampling temperature of 0.4. 
The balancing coefficient $\alpha$ for the rectification loss is set to 1.0. 
We optimize the model using AdamW with a learning rate of 5e-7. 
Unless otherwise specified, we inherit all experiment settings from previous work~\cite{wei2025streamvln,zhang2024navid,cheng2024navila}, and only fine-tune the LLM via LoRA~\cite{hu2022lora}.
The training process is conducted on NVIDIA H800 GPUs. BudVLN training requires only 27 GPU hours vs. 114 hours for DAgger.
All subsequent evaluations are performed on NVIDIA RTX 4090 GPUs.
The agent operates with strictly monocular RGB inputs.

\subsection{Main Results}
Table \ref{tab:combined_results} presents the quantitative comparison on the R2R-CE and RxR-CE benchmarks. 
BudVLN establishes a new state-of-the-art across both datasets.
Remarkably, despite relying exclusively on monocular RGB inputs, BudVLN outperforms not only strong RGB-based baselines but also competitive methods equipped with privileged sensors.

\noindent\textbf{Performance on R2R-CE.} 
Our method achieves the lowest Navigation Error (NE) of 4.74m and the highest Success Rate (SR) of 57.6\% among all compared methods on R2R-CE. Crucially, this efficiency is comparable to or even surpasses map-based methods, validating that our GRPO-driven optimality seeking effectively generates concise paths without requiring global maps.

\noindent\textbf{Performance on RxR-CE.} 
On the more challenging RxR-CE dataset, BudVLN demonstrates exceptional generalization capabilities.
It sets new records across all key metrics, achieving an NE of 5.79m, an SR of 56.1\%, an SPL of 46.6\%, and an nDTW of 63.2\%.
These results confirm the robustness of our dynamic guidance framework in handling complex instructions and long-horizon navigation.

\begin{figure*}[t]
\centering
\includegraphics[width=\linewidth]{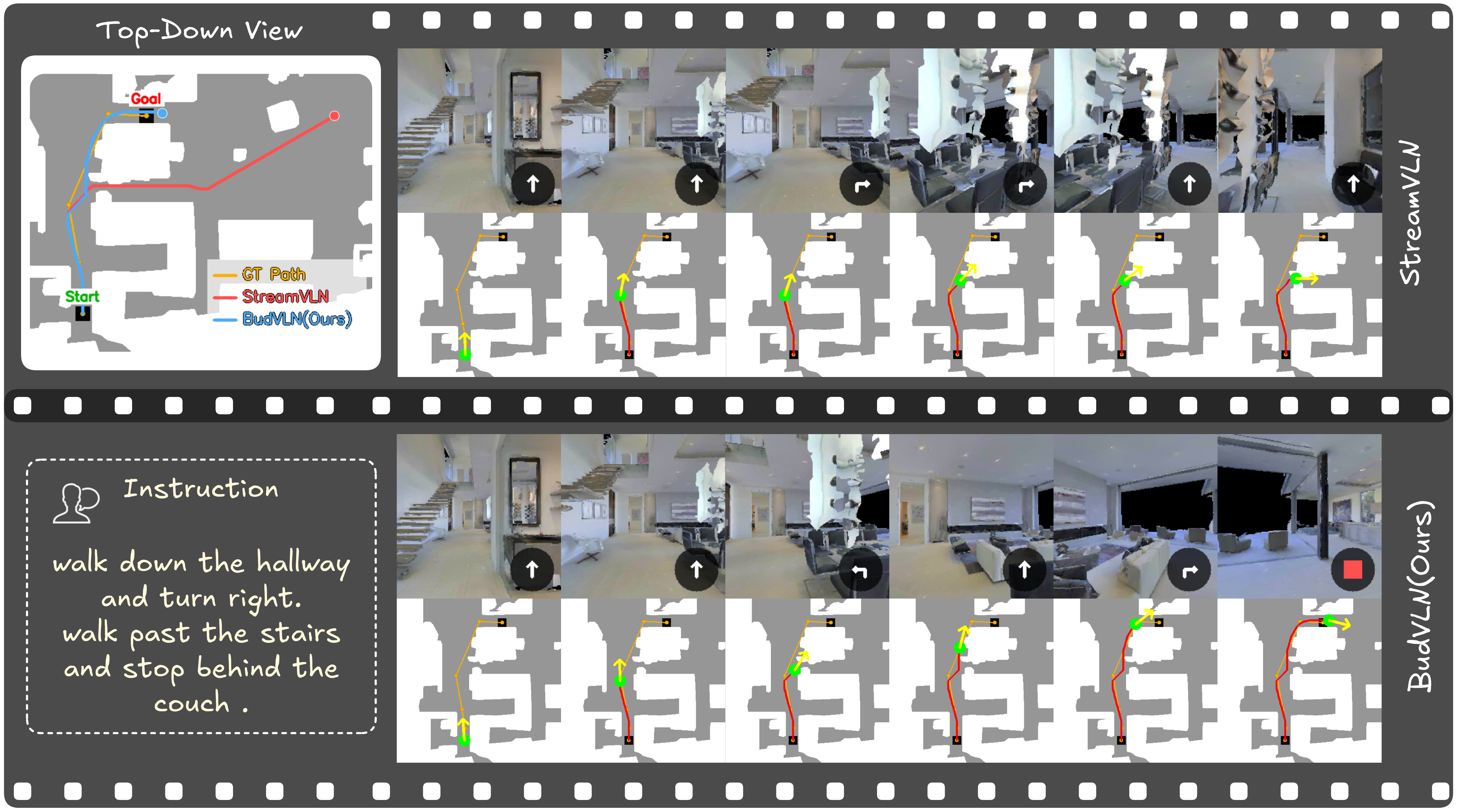} 
\caption{
\textbf{Qualitative comparison between the Baseline and BudVLN.} 
The top row illustrates the Baseline agent failing to ground the instruction, leading to a deviation from the correct path and eventually getting stuck. 
In contrast, the bottom row demonstrates that BudVLN, trained with our \textit{Retrospective Rectification} and \textit{GRPO}, successfully navigates the complex environment.
Despite visual similarities in the initial steps, BudVLN exhibits superior robustness, effectively avoiding the failure mode that trapped the baseline.
}
\label{fig:qualitative}
\end{figure*}

\subsection{Ablation Study}
\label{sec:ablation}

We conduct ablation studies on the R2R-CE \textit{Validation Unseen} split to investigate the individual contributions of each component and validate our design choices.

\noindent\textbf{Impact of Individual Components.} 
We verify the distinct roles of the Failure Prevention (Rectification) and Optimality Seeking by analyzing the incremental performance changes reported in Table \ref{tab:ablation_components}.

\begin{itemize}
    \item \textbf{Role of Rectification:} 
    Comparing the Baseline with the variant without GRPO, we assess the impact of adding the Rectification module alone. 
    Rectification serves as the primary driver for robustness, boosting the Success Rate (SR) from 57.0\% to 57.2\% and reducing Navigation Error (NE) from 4.87m to 4.82m. 
    Interestingly, adding Rectification alone slightly degrades SPL (50.5\% $\rightarrow$ 50.1\%) despite improving SR. 
    This suggests that while SFT effectively rescues hard failure cases, focusing solely on error correction introduces a distributional bias that inadvertently harms the execution efficiency of previously mastered instructions. 
    This highlights the necessity of the proficiency pathway: GRPO is crucial not just for optimization, but for preserving and reinforcing the efficiency of proficient samples against the gradient interference from corrective SFT updates.
    
    \item \textbf{Synergy with GRPO:} 
    The Full Model integrates the GRPO mechanism alongside rectification. 
    Crucially, GRPO reverses the decline in efficiency, significantly boosting SPL from 50.1\% to 51.1\%, while further enhancing SR to 57.6\% and reducing NE to 4.74m.
    This result confirms the synergy between the two components: Rectification acts as a safety net to ensure high completion rates (fixing hard samples), while GRPO optimizes the policy to traverse these paths as efficiently as possible, effectively mitigating the detour costs introduced by correction.
\end{itemize}

\begin{table}[h]
\centering
\resizebox{0.95\linewidth}{!}{
\begin{tabular}{l|cc|cccc}
\toprule
Model Variant & GRPO & Rect. &
OS $\uparrow$ & SR $\uparrow$ & SPL $\uparrow$ & NE $\downarrow$ \\
\midrule
Baseline & - & - &  64.3  & 57.0 & 50.5 & 4.87 \\
w/o GRPO & - & \checkmark &  64.0  & 57.2 & 50.1 & 4.82 \\ 
\textbf{BudVLN (Full)} & \checkmark & \checkmark &  \textbf{65.6}  & \textbf{57.6} & \textbf{51.1} & \textbf{4.74} \\
\bottomrule
\end{tabular}
}
\caption{Component analysis on R2R-CE Val Unseen.}
\label{tab:ablation_components}
\end{table}


\begin{table}[h]
\centering
\resizebox{0.85\linewidth}{!}{ 
\begin{tabular}{l|ccccc} 
\toprule
Strategy & OS $\uparrow$ & SR $\uparrow$ & SPL $\uparrow$ & NE $\downarrow$ & Time (h) $\downarrow$ \\
\midrule
Baseline & 64.3 & 57.0 & 50.5 & 4.87 & - \\ 
DAgger   & 64.3 & 57.1 & 50.7 & 4.87 & 114 \\ 
\textbf{Ours} & \textbf{65.6} & \textbf{57.6} & \textbf{51.1} & \textbf{4.74} & \textbf{27} \\
\bottomrule
\end{tabular}
}
\caption{Effectiveness of different rectification strategies.}
\label{tab:ablation_strategy}
\end{table}

\noindent\textbf{Effectiveness of Strategy.} 
A core contribution of this work is addressing the Instruction-State Misalignment problem during correction. In Table \ref{tab:ablation_strategy}, we compare our Dynamic Anchoring strategy (on-policy rollback to $s_{anc}$) against the standard DAgger approach (off-policy correction from the error state $s_{err}$).

Standard DAgger yields an SR of 57.1\%, which shows a marginal improvement over the baseline 57.0\% but trails behind our method 57.6\%. 
We attribute this limited gain to the semantic conflict inherent in DAgger: providing oracle actions at a deviated state $s_{err}$ often contradicts the agent's accumulated visual history, causing policy confusion. 
In contrast, our history-aware rollback ensures that supervision is provided on a valid path segment anchor. This maintains consistency between the visual history and the ground-truth action, leading to superior alignment and optimal performance SPL 51.1\% vs. 50.7\%.

\subsection{Qualitative Analysis}
\label{sec:qualitative}
To interpret the gains in Table \ref{tab:ablation_components}, Figure \ref{fig:qualitative} visualizes a representative episode where both agents follow the same instruction from the same start pose.
Baseline Failure. StreamVLN starts correctly but drifts at the intermediate stage, losing instruction-state alignment and stopping prematurely before reaching the target.
BudVLN Success. BudVLN remains robust at the critical turning point and reaches the goal, indicating that retrospective rectification improves on-policy correction and preserves history-action consistency.
Overall, the example supports our claim that dynamic on-policy guidance is crucial for long-horizon VLN.
\section{Conclusion}
We introduced BudVLN, a unified online framework to mitigate exposure bias and instruction-state misalignment in Vision-Language Navigation. Observing that DAgger-style corrections can induce semantic conflicts, we proposed retrospective rectification to synthesize on-policy supervision by re-anchoring to valid historical states. We further couple failure prevention with optimality seeking, using GRPO to refine proficient trajectories. Experiments show that BudVLN achieves state-of-the-art performance with only $\sim$25\% of the training cost of traditional methods. These results highlight the importance of strict semantic alignment for robust embodied navigation in continuous environments.

\bibliographystyle{named}
\bibliography{ijcai26}

\end{document}